%% file: ms.tex
\documentclass{article}
\usepackage{ijcai17}
\usepackage{times}
\usepackage{hyperref,url,booktabs,amsfonts,nicefrac,microtype,amsmath,amssymb,mathtools,xcolor,graphicx,float,fancyhdr}
\usepackage{natbib}
\newcommand{\norm}[1]{\left\lVert#1\right\rVert}
\newcommand{\inR}[2]{\in \mathbb{R}^{#1 \times #2}}
\DeclareMathOperator*{\argmin}{arg\,min}
\usepackage[skip=4pt]{caption}

\title{Right for the Right Reasons: Training Differentiable Models by Constraining their Explanations}

\author{Andrew Ross, Michael C. Hughes, and Finale Doshi-Velez \\
Paulson School of Engineering and Applied Sciences, Harvard University, Cambridge, MA 02138, USA  \\
andrew\_ross@g.harvard.edu, mchughes@seas.harvard.edu, finale@seas.harvard.edu}

\begin{document}

\maketitle

\begin{abstract}

Neural networks are among the most accurate supervised learning
methods in use today.  However, their opacity makes them difficult to
trust in critical applications, especially if conditions in training
may differ from those in test.  Recent work on explanations for
black-box models has produced tools (e.g. LIME) to show the implicit
rules behind predictions.  These tools can help us identify when
models are right for the wrong reasons. However, these methods do not
scale to explaining entire datasets and cannot correct the problems
they reveal. We introduce a method for efficiently explaining and
regularizing differentiable models by examining and selectively
penalizing their input gradients. We apply these penalties both based
on expert annotation and in an unsupervised fashion that produces
multiple classifiers with qualitatively different decision
boundaries. On multiple datasets, we show our approach generates
faithful explanations and models that generalize much better when
conditions differ between training and test.

\end{abstract}


\section{Introduction}

High-dimensional real-world datasets are often full of
ambiguities. When we train classifiers on such data, it is frequently
possible to achieve high accuracy using classifiers with qualitatively
different decision boundaries. To narrow down our choices and
encourage robustness, we usually employ regularization techniques
(e.g. encouraging sparsity or small parameter values).  We also
structure our models to ensure domain-specific invariances (e.g. using
convolutional neural nets when we would like the model to be invariant
to spatial transformations).  However, these solutions do not address
situations in which our training dataset contains subtle confounds or
differs qualitatively from our test dataset.  In these cases, our
model may fail to generalize no matter how well it is tuned.

Such generalization gaps are of particular concern for uninterpretable models
such as neural networks, especially in sensitive domains. For example,
\citet{caruana2015intelligible} describe a model intended to prioritize care
for patients with pneumonia.  The model was trained to predict hospital
readmission risk using a dataset containing attributes of patients hospitalized at least once for pneumonia.  Counterintuitively, the model learned that the
presence of asthma was a \textit{negative} predictor of readmission, when in
reality pneumonia patients with asthma are at a greater medical risk.  This model would have presented a grave safety risk if used in production.  This
problem occurred because the outcomes in the dataset reflected not just the
severity of patients' diseases but the quality of care they initially received,
which was higher for patients with asthma.

This case and others like it have motivated recent work in interpretable
machine learning, where algorithms provide explanations for domain experts to
inspect for correctness before trusting model predictions.  However, there has
been limited work in optimizing models to find not just the right prediction
but also the \textit{right explanation}.  Toward this end, this work makes the
following contributions:

\begin{itemize}
  \item We confirm empirically on several datasets that input gradient
    explanations match state of the art sample-based explanations (e.g. LIME
    \citep{limegithub}).
  \item Given annotations about incorrect explanations for particular
    inputs, we efficiently optimize the classifier to learn alternate
    explanations (to be right for better reasons).
  \item When annotations are not available, we sequentially discover
    classifiers with similar accuracies but qualitatively different
    decision boundaries for domain experts to inspect for validity.
\end{itemize}

\subsection{Related Work}

We first define several important terms in interpretable machine learning.  All
classifiers have \textit{implicit decision rules} for converting an input into
a decision, though these rules may be opaque.  A model is
\textit{interpretable} if it provides explanations for its predictions in a
form humans can understand; an \textit{explanation} provides reliable
information about the model's implicit decision rules for a given prediction.
In contrast, we say a machine learning model is \textit{accurate} if most of
its predictions are correct, but only \textit{right for the right reasons} if
the implicit rules it has learned generalize well and conform to domain
experts' knowledge about the problem.

Explanations can take many forms \citep{explanations} and evaluating
the quality of explanations or the interpretability of a model is
difficult \citep{interpretability-mythos,rigorous}. However, within the machine
learning community recently there has been convergence
\citep{unexpected-unity} around local counterfactual explanations,
where we show how perturbing an input $x$ in various ways will affect
the model's prediction $\hat{y}$. This approach to explanations can be
domain- and model-specific (e.g. ``annotator rationales'' used to
explain text classifications in
\citet{erasure,rationalizing,rationalenetworks}).
Alternatively, explanations can be model-agnostic and
relatively domain-general, as exemplified by LIME (Local Interpretable
Model-agnostic Explanations, \citep{lime,programs-as-explanations}) which
trains and presents local sparse models of how predictions change
when inputs are perturbed.

\begin{figure}[b]
\begin{center}
\includegraphics[width=0.49\textwidth]{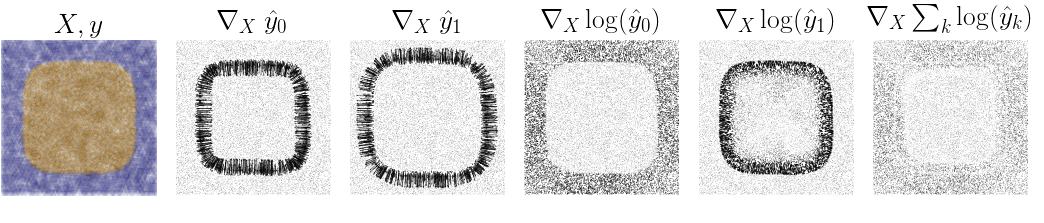} \\
\includegraphics[width=0.49\textwidth]{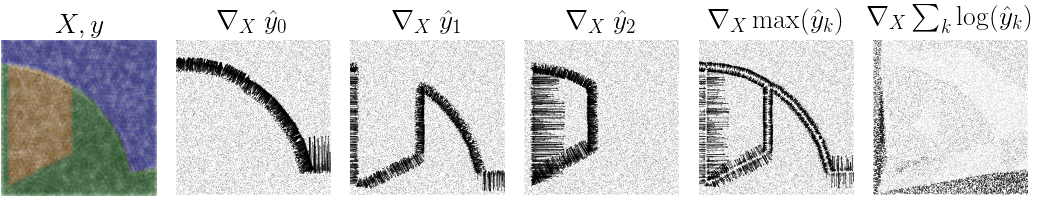}
\end{center}
\caption{Input gradients lie normal to the model's decision boundary. Examples
  above are for simple, 2D, two- and three-class datasets, with input gradients
  taken with respect to a two hidden layer multilayer perceptron with ReLU
  activations. Probability input gradients are sharpest near decision
boundaries, while log probabilities input gradients are more consistent within
decision regions. The sum of log probability gradients contains information
about the full model.}
\label{fig:simple-2d-explanations}
\end{figure}

The per-example perturbing and fitting process used in models such as LIME can be
computationally prohibitive, especially if we seek to explain an entire dataset
during each training iteration.  If the underlying model is differentiable,
one alternative is to use input gradients as local explanations
(\citet{baehrens2010explain} provides a particularly good introduction; see
also
\citet{grad-cam,convnets-input-gradients,text-input-gradients,input-gradients2}).
The idea is simple: the gradients of the model's output probabilities with
respect to its inputs literally describe the model's decision boundary (see
Figure~\ref{fig:simple-2d-explanations}). They are similar in spirit to the
local linear explanations of LIME but much faster to compute.

Input gradient explanations are not perfect for all use-cases---for points far
from the decision boundary, they can be uniformatively small and
do not always capture the idea of salience (see discussion and alternatives
proposed in
\citet{deeplift,layerwise,deeptaylor,grad-counterfactuals,shrikumar2017learning,fong2017interpretable}).
However, they are exactly what is required for constraining the decision
boundary.  In the past, \citet{doublebackprop} showed that applying penalties
to input gradient magnitudes can improve generalization; to our knowledge, our
application of input gradients to constrain explanations and find alternate
explanations is novel.

More broadly, none of the works above on interpretable machine
learning attempt to optimize explanations for correctness.  For SVMs
and specific text classification architectures, there exists work on
incorporating human input into decision boundaries in the form of
annotator rationales \citep{zaidan,donahue,rationalenetworks}.
Unlike our approach, these works are either tailored to specific domains
or do not fully close the loop between generating explanations and
constraining them.

\subsection{Background: Input gradient explanations}

Consider a differentiable model $f$ parametrized by $\theta$ with
inputs $X \inR{N}{D}$ and probability vector outputs $f(X|\theta) =
\hat{y} \inR{N}{K}$ corresponding to one-hot labels $y
\inR{N}{K}$. Its \textit{input gradient} is given by $f_X(X_n|\theta)$ or
$\nabla_X \hat{y}_n,$ which is a vector normal to the model's decision boundary
at $X_n$ and thus serves as a first-order description of the model's
behavior near $X_n$. The gradient has the same shape as each vector
$X_n$; large-magnitude values of the input gradient
indicate elements of $X_n$ that would affect $\hat{y}$ if changed.  We
can visualize explanations by highlighting portions of $X_n$ in
locations with high input gradient magnitudes.


\section{Our Approach}

We wish to develop a method to train models that are right for the right
reasons. If explanations faithfully describe a model's underlying behavior,
then constraining its explanations to match domain knowledge should cause its
underlying behavior to more closely match that knowledge too. We first describe
how input gradient-based explanations lend themselves to efficient optimization
for correct explanations in the presence of domain knowledge, and then describe
how they can be used to efficiently search for qualitatively different decision
boundaries when such knowledge is not available.

\subsection{Constraining explanations in the loss function} \label{sec:loss}

When constraining input gradient explanations, there are two basic options: we
can either constrain them to be large in relevant areas or small in irrelevant
areas. However, because input gradients for relevant inputs in many models
\textit{should} be small far from the decision boundary, and because we do not
know in advance how large they should be, we opt to shrink irrelevant gradients
instead.

Formally, we define an annotation matrix $A \in \{0,1\}^{N \times D}$, which
are binary masks indicating whether dimension $d$ should be irrelevant for
predicting observation $n$. We would like $\nabla_X \hat{y}$ to be near
$0$ at these locations.  To that end, we optimize a loss function $L(\theta, X,
y, A)$ of the form \[
\begin{split}
  & L(\theta, X, y, A) = \underbrace{\sum_{n=1}^N\sum_{k=1}^K -y_{nk} \log(\hat{y}_{nk})}_{\text{Right answers}} \\ & + \underbrace{\lambda_1\sum_{n=1}^N\sum_{d=1}^D \left(A_{nd} \frac{\partial}{\partial x_{nd}} \sum_{k=1}^K  \log(\hat{y}_{nk}) \right)^2}_{\text{Right reasons}} + \underbrace{\lambda_2 \sum_{i} \theta_i^2}_{\text{Regular}},
\end{split}
\]
which contains familiar cross entropy and $\theta$ regularization terms along
with a new regularization term that discourages the input gradient from being
large in regions marked by $A$. This term has a regularization parameter
$\lambda_1$ which should be set such that the ``right answers'' and ``right
reasons'' terms have similar orders of magnitude; see Appendix
\ref{sec:crossval} for more details. 
Note that this loss penalizes the gradient of the
\textit{log} probability, which performed best in
practice, though in many visualizations we show $f_X$, which is the gradient of the predicted probability itself. Summing across classes led to slightly more stable results than
using the predicted class log probability $\max \log(\hat{y}_k)$, perhaps due
to discontinuities near the decision boundary (though both methods were
comparable). We did not explore regularizing input gradients of specific class
probabilities, though this would be a natural extension.

Because this loss function is differentiable with respect to $\theta$, we can
easily optimize it with gradient-based optimization methods.  We do not need
annotations (nonzero $A_n$) for every input in $X$, and in the case $A =
0^{N \times D}$, the explanation term has no effect on the loss. At the other
extreme, when $A$ is a matrix of all 1s, it encourages the model to have small
gradients with respect to its inputs; this can improve generalization on its own
\citep{doublebackprop}. Between those extremes, it biases our model against
\textit{particular} implicit rules.

This penalization approach enjoys several desirable properties.
Alternatives that specify a single $A_{d}$ for all examples presuppose
a coherent notion of global feature importance, but when
decision boundaries are nonlinear many features are only relevant in the
context of specific examples.  Alternatives that simulate
perturbations to entries known to be irrelevant (or to determine
relevance as in \citet{lime}) require defining domain-specific
perturbation logic; our approach does not.  Alternatives that apply
hard constraints or completely remove elements identified by
$A_{nd}$ miss the fact that the entries in $A$ may be
imprecise even if they are human-provided.  Thus, we opt to preserve
potentially misleading features but softly penalize their use.

\subsection{Find-another-explanation: discovering many possible rules without annotations} \label{sec:fae}

Although we can obtain the annotations $A$ via experts as in \citet{zaidan}, we
may not always have this extra information or know the ``right reasons.'' In
these cases, we propose an approach that iteratively adapts $A$ to discover
multiple models accurate for \emph{qualitatively different} reasons; a domain
expert could then examine them to determine which is the right for the best
reasons.  Specifically, we generate a ``spectrum'' of models with different
decision boundaries by iteratively training models, explaining $X$, then
training the next model to differ
from previous iterations: \[
\begin{aligned}
  A_0 & = 0,                                      & \theta_0 = \argmin_{\theta} L(\theta, X, y, A_0), \\
  A_1 & = M_c\left[f_X|\theta_0\right],           & \theta_1 = \argmin_{\theta} L(\theta, X, y, A_1), \\
A_2 & = M_c\left[f_X|\theta_1\right] \cup A_1,\ & \theta_2 = \argmin_{\theta} L(\theta, X, y, A_2), \\
\end{aligned}
\]
\noindent \begin{center}$\hdots$\end{center}
where the function $M_c$ returns a binary mask indicating which gradient
components have a magnitude ratio (their magnitude divided by the largest
component magnitude) of at least $c$ and where we abbreviated the
input gradients of the entire training set $X$ at $\theta_i$ as
$f_X|\theta_i$. In other words, we regularize input gradients where they were
largest in magnitude
previously.  If, after repeated iterations, accuracy decreases or
explanations stop changing (or only change after significantly increasing
$\lambda_1$), then we have spanned the space of possible models.  All of the
resulting models will be accurate, but for different reasons; although we do
not know which reasons are best, we can present them to a domain expert for
inspection and selection. We can also prioritize labeling or reviewing examples
about which the ensemble disagrees. Finally, the size of the ensemble
provides a rough measure of dataset redundancy.

\section{Empirical Evaluation}

We demonstrate explanation generation, explanation constraints, and the
find-another-explanation method on a toy color dataset and three real-world
datasets. In all cases, we used a multilayer perceptron with two hidden layers
of size 50 and 30, ReLU nonlinearities with a softmax output, and a $\lambda_2
= 0.0001$ penalty on $\norm{\theta}_2^2$.  We trained the network using Adam
\citep{adam} (with a batch size of 256) and Autograd \citep{autograd}.  For
most experiments, we used an explanation L2 penalty of $\lambda_1 = 1000$,
which gave our ``right answers'' and ``right reasons'' loss terms similar
magnitudes. More details about cross-validation are included in
Appendix~\ref{sec:crossval}.  For the cutoff value $c$ described in Section
\ref{sec:fae} and used for display, we often chose 0.67, which tended to
preserve 2-5\% of gradient components (the average number of qualifying
elements tended to fall exponentially with $c$).  Code for all experiments is
available at
\texttt{\href{https://github.com/dtak/rrr}{https://github.com/dtak/rrr}}.


\subsection{Toy Color Dataset}

We created a toy dataset of $5 \times 5 \times 3$ RGB images with four possible
colors.  Images fell into two classes with two independent
decision rules a model could implicitly learn: whether their four corner pixels
were all the same color, and whether their top-middle three pixels were all different
colors. Images in class 1 satisfied both conditions and images in class 2
satisfied neither.  Because only corner and top-row pixels are relevant, we
expect any faithful explanation of an accurate model to highlight them.

\begin{figure}[b]
  \centering
  \includegraphics[width=0.33\textwidth]{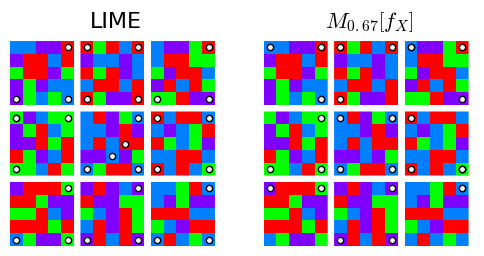}
  \caption{Gradient vs. LIME explanations of nine perceptron predictions on the
    Toy Color dataset. For gradients, we plot dots above
    pixels identified by $M_{0.67}\left[f_X\right]$ (the top 33\%
  largest-magnitude input gradients), and for LIME, we select the top 6
features (up to 3 can reside in the same RGB pixel). Both methods suggest that
the model learns the corner rule.}
  \label{fig:colors-vs-lime}
\end{figure}

In Figure \ref{fig:colors-vs-lime}, we see both LIME and input
gradients identify the same relevant pixels, which suggests that (1)
both methods are effective at explaining model predictions, and (2)
the model has learned the corner rather than the top-middle rule,
which it did consistently across random restarts.

\begin{figure}[t]
\centering
\includegraphics[width=0.49\textwidth]{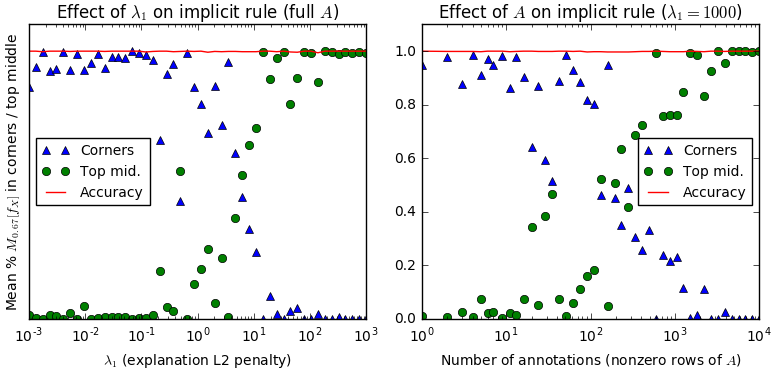}
\caption{Implicit rule transitions as we increase $\lambda_1$ and the number of
  nonzero rows of $A$. Pairs of points represent the fraction of
  large-magnitude ($c=0.67$) gradient components in the corners and top-middle
  for 1000 test examples, which almost always add to 1 (indicating the
  model is most sensitive to these elements alone, even during
transitions). Note there is a wide regime where the model learns a
hybrid of both rules.}
\label{fig:color-transitions}
\end{figure}

\begin{figure}[t]
\centering
\includegraphics[width=0.49\textwidth]{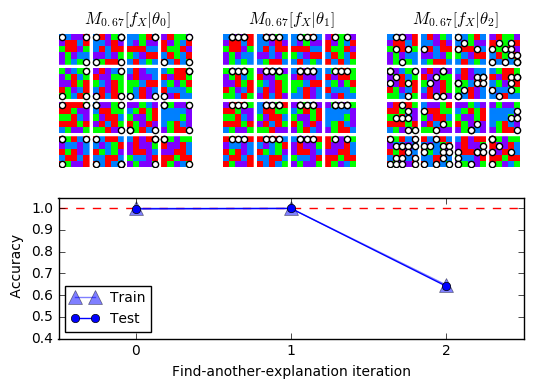}
\caption{Rule discovery using find-another-explanation method with 0.67 cutoff
and $\lambda_1=10^3$ for $\theta_1$ and $\lambda_1=10^6$ for $\theta_2$. Note
how the first two iterations produce explanations corresponding to the two
rules in the dataset while the third produces very noisy explanations (with low
accuracies).}
\label{fig:color-fae}
\end{figure}

However, if we train our model with a nonzero $A$ (specifically, setting
$A_{nd}=1$ for corners $d$ across examples $n$), we were able to cause it to
use the other rule.  Figure \ref{fig:color-transitions} shows how the model
transitions between rules as we vary $\lambda_1$ and the number of examples
penalized by $A$.  This result demonstrates that the model can be made to learn
multiple rules despite only one being commonly reached via standard
gradient-based optimization methods. However, it depends on knowing a good
setting for $A$, which in this case would still require annotating on the order
of $10^3$ examples, or 5\% of our dataset (although always including examples
with annotations in Adam minibatches let us consistently switch rules with
only 50 examples, or 0.2\% of the dataset).

Finally, Figure \ref{fig:color-fae} shows we can use the
find-another-explanation technique from Sec.~\ref{sec:fae} to discover the
other rule without being given $A$.  Because only two rules lead to high
accuracy on the test set, the model performs no better than random guessing
when prevented from using either one (although we have to increase the penalty
high enough that this accuracy number may be misleading - the essential point
is that after the first iteration, explanations stop changing).  Lastly, though
not directly relevant to the discussion on interpretability and explanation, we
demonstrate the potential of explanations to reduce the amount of data required
for training in Appendix~\ref{sec:learnfast}.


\subsection{Real-world Datasets}

To demonstrate real-world, cross-domain applicability, we test our approach on
variants of three familiar machine learning text, image, and tabular datasets:

\begin{itemize}
\item \textbf{20 Newsgroups:} As in \citet{lime}, we test input
  gradients on the \texttt{alt.atheism}
  vs. \texttt{soc.religion.christian} subset of the 20 Newsgroups
  dataset \citep{uciml}. We used the same two-hidden layer network
  architecture with a TF-IDF vectorizer with 5000 components, which
  gave us a 94\% accurate model for $A=0$.

\item \textbf{Iris-Cancer:} We concatenated all examples in classes 1 and 2
  from the Iris dataset with the the first 50 examples from each class in the
  Breast Cancer Wisconsin dataset \citep{uciml} to create a composite dataset
  $X\inR{100}{34},y\in\{0,1\}$. Despite the dataset's small size, our network
  still obtains an average test accuracy of 92\% across 350 random
  $\frac{2}{3}$-$\frac{1}{3}$ training-test splits.  However, when we modify
  our test set to remove the 4 Iris components, average test accuracy falls to
  81\% with higher variance, suggesting the model learns to depend on Iris
  features and suffers without them.  We verify that our explanations reveal
  this dependency and that regularizing them avoids it.

\item \textbf{Decoy MNIST:} On the baseline MNST dataset \citep{mnist}, our
  network obtains 98\% train and 96\% test accuracy. However, in Decoy MNIST,
  images $x$ have $4 \times 4$ gray swatches in randomly chosen corners whose
  shades are functions of their digits $y$ in training (in particular, $255 -
  25y$) but are random in test. On this dataset, our model has a higher 99.6\%
  train accuracy but a much lower 55\% test accuracy, indicating that the decoy
  rule misleads it. We verify that both gradient and LIME explanations let
  users detect this issue and that explanation regularization lets us
  overcome it.
\end{itemize}

\begin{figure*}[t]
	\begin{center}
	\includegraphics[width=0.85\textwidth]{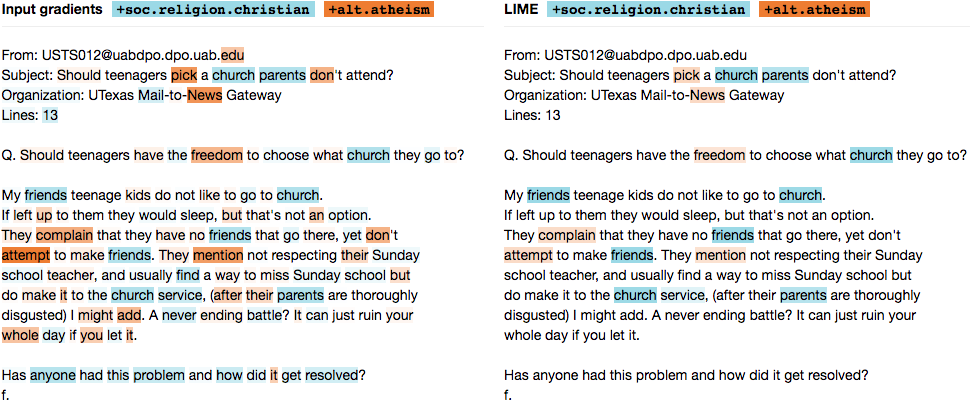}
	\end{center}
  \caption{Words identified by LIME vs. gradients on an example from the atheism vs.
    Christianity subset of 20 Newsgroups. More
    examples are available at
    \texttt{\href{https://github.com/dtak/rrr}{https://github.com/dtak/rrr}}.
    Words are blue if they support \texttt{soc.religion.christian}
    and orange if they support \texttt{alt.atheism}, with opacity
    equal to the ratio of the magnitude of the word's
    weight to the largest magnitude weight. LIME generates sparser explanations
    but the weights and signs of terms identified by both methods match
    closely. Note that both methods reveal some aspects of the model that
    are intuitive (``church'' and ``service'' are associated with Christianity), some aspects that are not (``13'' is associated with
    Christianity, ``edu'' with atheism), and some that are debatable (``freedom''
    is associated with atheism, ``friends'' with Christianity).}
	\label{fig:20ng-vs-lime}
\end{figure*}

\begin{figure}[b]
	\begin{center}
	\includegraphics[width=0.49\textwidth]{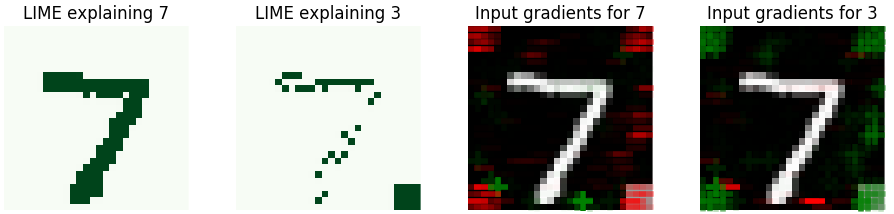}
	\end{center}
  \caption{Input gradient explanations for Decoy MNIST vs. LIME, using the LIME
  image library \citep{limegithub}. In this example, the
  model incorrectly predicts 3 rather than 7 because of the decoy swatch.}
	\label{fig:mnist-vs-lime}
\end{figure}

\begin{figure}[b]
	\begin{center}
	\includegraphics[width=0.49\textwidth]{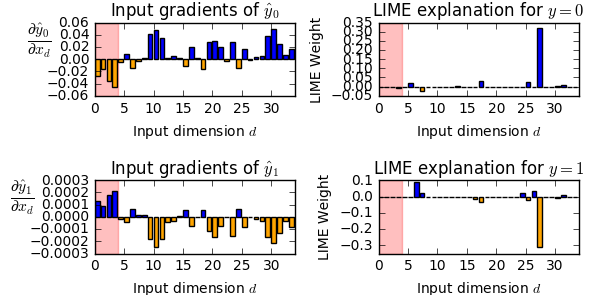}
	\end{center}
  \caption{Iris-Cancer features identified by input gradients vs. LIME, with
  Iris features highlighted in red. Input gradient explanations are more
faithful to the model. Note that most gradients change sign
when switching between $\hat{y}_0$ and $\hat{y}_1$, and that
the magnitudes of input gradients are different across examples,
which provides information about examples' proximity to the decision boundary.}
	\label{fig:iris-vs-lime}
\end{figure}

\noindent \textbf{Input gradients are consistent with sample-based methods such as
LIME, and faster.} On 20 Newsgroups (Figure \ref{fig:20ng-vs-lime}), input
gradients are less sparse but identify all of the same words in the document
with similar weights. Note that input gradients also identify words outside the
document that would affect the prediction if added.

\begin{figure}[b]
	\begin{center}
	\includegraphics[width=0.49\textwidth]{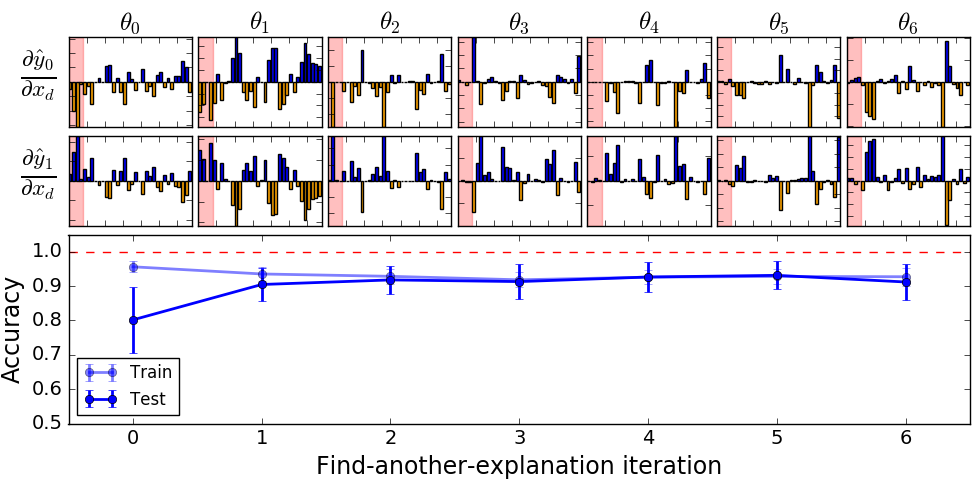}
	\includegraphics[width=0.49\textwidth]{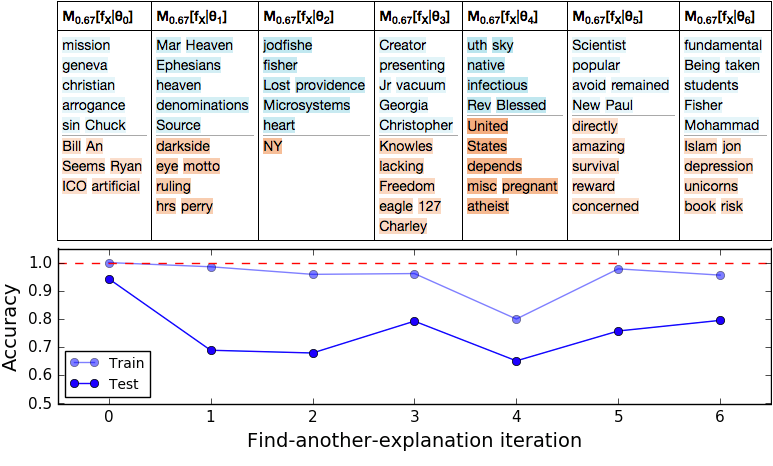}
	\includegraphics[width=0.49\textwidth]{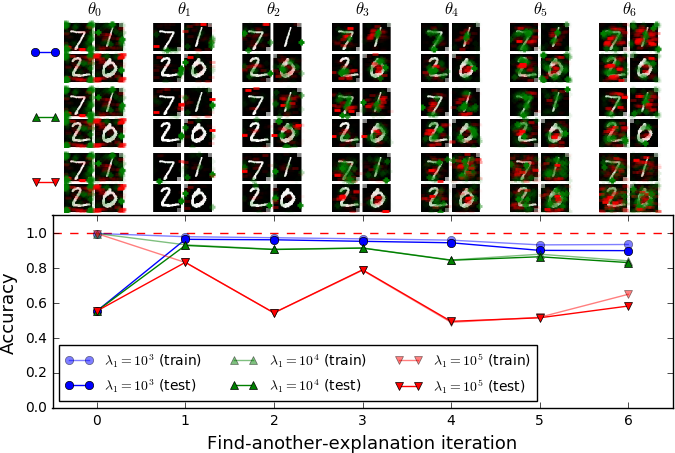}
	\end{center}
  \caption{Find-another-explanation results on Iris-Cancer (top; errorbars show
    standard deviations across 50 trials), 20 Newsgroups (middle; blue supports
    Christianity and orange supports atheism, word opacity set to magnitude
    ratio), and Decoy MNIST (bottom, for three values of $\lambda_1$ with
    scatter opacity set to magnitude ratio cubed). Real-world datasets are
    often highly redundant and allow for diverse models with similar
  accuracies. On Iris-Cancer and Decoy MNIST, both explanations and accuracy
results indicate we overcome confounds after 1-2 iterations without any prior
knowledge about them encoded in $A$.}
	\label{fig:fae}
\end{figure}

On Decoy MNIST (Figure \ref{fig:mnist-vs-lime}), both LIME and input gradients
reveal that the model predicts 3 rather than 7 due to the color
swatch in the corner. Because of their fine-grained resolution, input gradients
sometimes better capture counterfactual behavior, where extending or adding
lines outside of the digit to either reinforce it or transform it into another
digit would change the predicted probability (see also Figure \ref{fig:fae}).
LIME, on the other hand, better captures the fact that the main portion of the
digit is salient (because its super-pixel perturbations add and remove larger
chunks of the digit).

On Iris-Cancer (Figure \ref{fig:iris-vs-lime}), input gradients actually
outperform LIME.  We know from the accuracy difference that Iris features are
important to the model's prediction, but LIME only identifies a single
important feature, which is from the Breast Cancer dataset (even when we vary
its perturbation strategy).  This example, which is tabular and contains
continuously valued rather categorical features, may represent a pathological
case for LIME, which operates best when it can selectively mask a small number
of meaningful chunks of its inputs to generate perturbed samples.  For truly
continuous inputs, it should not be surprising that explanations based on
gradients perform best.

There are a few other advantages input gradients have over
sample-based perturbation methods. On 20 Newsgroups, we noticed that
for very long documents, explanations generated by the sample-based
method LIME are often overly sparse (see Appendix~\ref{sec:long20ng}),
and there are many words identified as significant by input gradients
that LIME ignores. This may be because the number of features LIME
selects must be passed in as a parameter beforehand, and it may also
be because LIME only samples a fixed number of times. For sufficiently
long documents, it is unlikely that sample-based approaches will mask
every word even once, meaning that the output becomes increasingly
nondeterministic---an undesirable quality for explanations. To resolve
this issue, one could increase the number of samples, but that would
increase the computational cost since the model must be evalutated at
least once per sample to fit a local surrogate.  Input gradients, on
the other hand, only require on the order of one model evaluation
\textit{total} to generate an explanation of similar quality
(generating gradients is similar in complexity to predicting
probabilities), and furthermore, this complexity is based on the
vector length, \textit{not} the document length. This issue
(underscored by Table \ref{tab:lime-vs-grad-performance}) highlights
some inherent scalability advantages input gradients enjoy over
sample-based perturbation methods.

\begin{table}[b]
\centering
\begin{tabular}{|c|c|c|c|} \hline
    & LIME & Gradients & Dimension of $x$ \\
\hline
Iris-Cancer & 0.03s & 0.000019s & 34 \\
\hline
Toy Colors    & 1.03s  & 0.000013s & 75 \\
\hline
Decoy MNIST & 1.54s  & 0.000045s & 784 \\
\hline
20 Newsgroups & 2.59s  & 0.000520s & 5000 \\
\hline
\end{tabular}
\caption{Gradient vs. LIME runtimes per explanation. Note that each method uses
  a different version of LIME; Iris-Cancer and Toy Colors use
  \texttt{lime\_tabular} with continuous and quartile-discrete perturbation
  methods, respectively, Decoy MNIST uses \texttt{lime\_image}, and 20
Newsgroups uses \texttt{lime\_text}. Code was executed on a laptop and input
gradient calculations were not optimized for performance, so runtimes are only
meant to provide a sense of scale.}

\label{tab:lime-vs-grad-performance}
\end{table}

\begin{figure}[b]
	\begin{center}
	\includegraphics[width=0.49\textwidth]{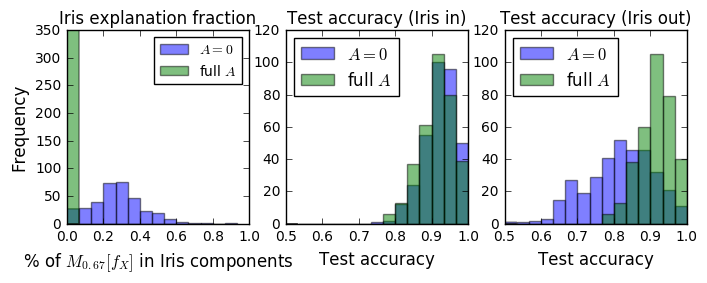}
  \end{center} \caption{Overcoming confounds using explanation constraints on
    Iris-Cancer (over 350 random train-test splits).  By default ($A=0$), input
    gradients tend to be large in Iris dimensions, which results in lower
    accuracy when Iris is removed from the test set. Models trained with
  $A_{nd}=1$ in Iris dimensions (full $A$) have almost exactly the same test
accuracy with and without Iris.}
	\label{fig:iris-confounds}
\end{figure}

\begin{figure}[b]
	\begin{center}
	\includegraphics[width=0.33\textwidth]{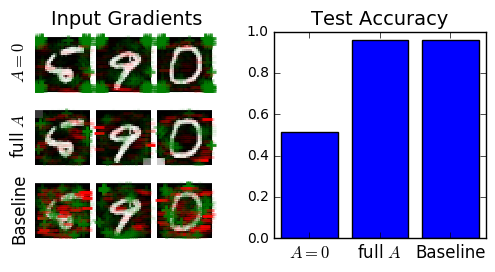}
	\end{center}
  \caption{Training with explanation constraints on Decoy MNIST. Accuracy is
    low ($A=0$) on the swatch color-randomized test set unless
    the model is trained with $A_{nd}=1$ in swatches (full $A$).
    In that case, test accuracy matches the same architecture's performance on
    the standard MNIST dataset (baseline).}
	\label{fig:mnist-confounds}
\end{figure}

$\,$ \\ \noindent \textbf{Given annotations, input gradient regularization finds solutions
consistent with domain knowledge.} Another key advantage of using an
explanation method more closely related to our model is that we can then
incorporate explanations into our training process, which are most useful when
the model faces ambiguities in how to classify inputs. We deliberately
constructed the Decoy MNIST and Iris-Cancer datasets to have this kind of
ambiguity, where a rule that works in training will not generalize to test.
When we train our network on these confounded datasets, their test accuracy is
better than random guessing, in part because the decoy rules are not simple and
the primary rules not complex, but their performance is still significantly
worse than on a baseline test set with no decoy rules. By penalizing
explanations we know to be incorrect using the loss function defined in Section
\ref{sec:loss}, we are able to recover that baseline test accuracy, which we
demonstrate in Figures \ref{fig:iris-confounds} and \ref{fig:mnist-confounds}.

$\,$ \\ \noindent \textbf{When annotations are unavailable, our
find-another-explanation method discovers diverse classifiers.} As we saw with
the Toy Color dataset, even if almost every row of $A$ is 0, we can still
benefit from explanation regularization (meaning practitioners can gradually
incorporate these penalties into their existing models without much upfront
investment).  However, annotation is never free, and in some cases we either do
not know the right explanation or cannot easily encode it.  Additionally, we
may be interested in exploring the structure of our model and dataset in a less
supervised fashion.  On real-world datasets, which are usually overdetermined,
we can use find-another-explanation to discover $\theta$s in shallower local
minima that we would normally never explore. Given enough models right for
different reasons, hopefully at least one is right for the right reasons.

Figure \ref{fig:fae} shows find-another-explanation results for our three
real-world datasets, with example explanations at each iteration above and
model train and test accuracy below. For Iris-Cancer, we find that the initial
iteration of the model heavily relies on the Iris features and has high train
but low test accuracy, while subsequent iterations have lower train
but higher test accuracy (with smaller gradients in Iris components). In other
words, we spontaneously obtain a more generalizable model without a predefined $A$
alerting us that the first four features are misleading.

Find-another-explanation also overcomes confounds on Decoy MNIST, needing only
one iteration to recover baseline accuracy. Bumping $\lambda_1$ too high (to
the point where its term is a few orders of magnitude larger than the
cross-entropy) results in more erratic behavior. Interestingly, in a process
remniscent of distillation \citep{distillation}, the gradients themselves
become more evenly and intuitively distributed at later iterations. In many
cases they indicate that the probabilities of certain digits increase when we
brighten pixels along or extend their distinctive strokes, and that they
decrease if we fill in unrelated dark areas, which seems desirable. However, by
the last iteration, we start to revert to using decoy swatches in some cases.

On 20 Newsgroups, the words most associated with \texttt{alt.atheism} and \texttt{soc.religion.christian} change between iterations but remain mostly intuitive in their associations.
Train accuracy mostly remains high while test accuracy is unstable.

For all of these examples, accuracy remains high even as decision boundaries
shift significantly. This may because real-world data tends to contain
significant redundancies.

\subsection{Limitations}

Input gradients provide faithful information about a model's rationale for a
prediction but trade interpretability for efficiency. In
particular, when input features are not individually meaningful to users (e.g.
for individual pixels or word2vec components), input gradients may be difficult to
interpret and $A$ may be difficult to specify. Additionally, because they can
be 0 far from the decision boundary, they do not capture the idea of salience
as well as other methods
\citep{visualizing-convnets,grad-counterfactuals,deeptaylor,layerwise,deeplift}.
However, they are necessarily faithful to the model and easy to incorporate
into its loss function. Input gradients are first-order linear approximations
of the model; we might call them first-order explanations.

\section{Conclusions and Future Work}

We have demonstrated that training models with input gradient penalties makes
it possible to learn generalizable decision logic even when our dataset
contains inherent ambiguities. Input gradients are consistent with sample-based
methods such as LIME but faster to compute and sometimes more faithful to the
model, especially when our inputs are continous. Our find-another-explanation
method can present a range of qualitatively different classifiers when such
detailed annotations are not available, which may be useful in practice if we
suspect each model is only right for the right reasons in certain regions.  Our
consistent results on several diverse datasets show that input gradients merit
further investigation as scalable tools for optimizable explanations; there
exist many options for further advancements such as weighted annotations $A$,
different penalty norms (e.g. L1 regularization to encourage sparse gradients),
and more general specifications of whether features should be positively or
negatively predictive of specific classes for specific inputs.

Finally, our ``right for the right reasons'' approach may be of use in solving
related problems, e.g. in maintaining robustness despite the presence of
adversarial examples \citep{distillation}, or seeing whether explanations and
explanation constraints can further the goals of fairness, accountability, and
transparency in machine learning (either by detecting indirect influence
\citep{auditing} or by constraining models to avoid it
\citep{fairness,fairnessimpact}).  Building on our find-another-explanation
results, another promising direction is to include humans in the loop to
interactively guide models towards correct explanations.  Overall, we feel that
developing methods of ensuring that models are right for better reasons is
essential to overcoming the inherent obstacles to generalization posed by
ambiguities in real-world datasets.

\paragraph{Acknowledgements}
FDV acknowledges support from DARPA W911NF-16-1-0561 and AFOSR
FA9550-17-1-0155, and MCH acknowledges support from Oracle Labs.  All
authors thank Arjumand Masood, Sam Gershman, Paul Raccuglia, Mali
Akmanalp, and the Harvard DTaK group for many helpful discussions and
insights.

\bibliographystyle{named}
\small
\bibliography{bibliography}

\pagebreak

\begin{appendix}
\input{supplement}
\end{appendix}

\end{document}

%% file: supplement.tex
\section{Cross-Validation}
\label{sec:crossval}

Most regularization parameters are selected to maximize accuracy on a
validation set. However, when your training and validation sets share the same
misleading confounds, validation accuracy may not be a good proxy for test
accuracy. Instead, we recommend increasing the explanation regularization
strength $\lambda_1$ until the cross-entropy and ``right reasons'' terms have
roughly equal magnitudes (which corresponds to the region of highest test
accuracy below). Intuitively, balancing the terms in this way should push our
optimization away from cross-entropy minima that violate the explanation
constraints specified in $A$ and towards ones that correspond to ``better
reasons.'' Increasing $\lambda_1$ too much makes the cross-entropy term
negligible. In that case, our model performs no better than random guessing.

\begin{figure}[htb]
\begin{center}
\includegraphics[width=0.38\textwidth]{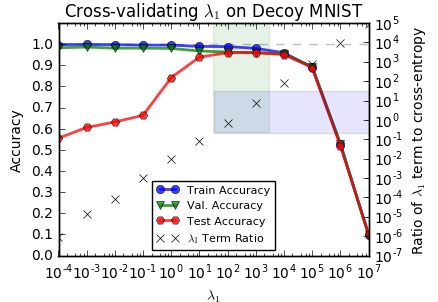}
\end{center}
\caption{\small Cross-validating $\lambda_1$. The regime of highest accuracy (highlighted) is also where the initial cross-entropy and $\lambda_1$ loss terms have similar magnitudes. Exact equality is not required; being an order of magnitude off does not significantly affect accuracy.}
\label{fig:crossval}
\end{figure}

\vspace{-0.5cm}
\section{Learning with Less Data} \label{sec:learnfast}

It is natural to ask whether explanations can reduce data requirements. Here we
explore that question on the Toy Color dataset using four variants of $A$ 
(with $\lambda_1$ chosen to match loss terms at each $N$).

\begin{figure}[htb]
\begin{center}
\includegraphics[width=0.48\textwidth]{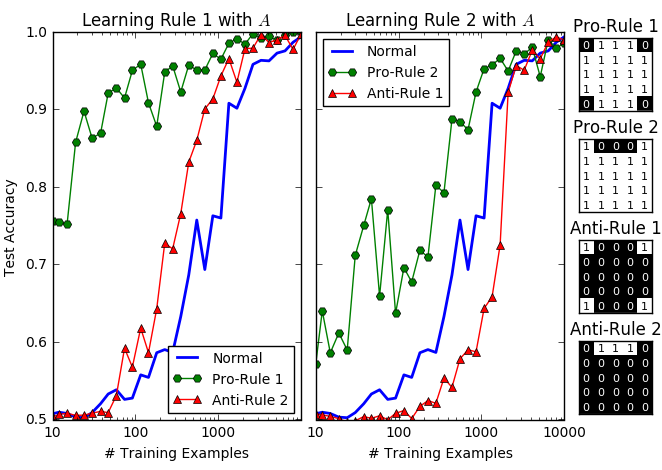}
\end{center}
  \caption{\small Explanation regularization can reduce data requirements.}
\label{fig:learnfast}
\end{figure}

\noindent We find that when $A$ is set to the Pro-Rule 1 mask, which penalizes
all pixels except the corners, we reach 95\% accuracy with fewer than 100
examples (as compared to $A=0$, where we need almost 10000). Penalizing the
top-middle pixels (Anti-Rule 2) or all pixels except the top-middle (Pro-Rule
2) also consistently improves accuracy relative to data. Penalizing the corners
(Anti-Rule 1), however, reduces accuracy until we reach a threshold $N$. This
may be because the corner pixels can match in 4 ways, while the
top-middle pixels can differ in $4\cdot3\cdot2=24$ ways, suggesting that Rule 2
could be inherently harder to learn from data and positional explanations
alone.

\section{Longer 20 Newsgroups Examples}
\label{sec:long20ng}
\vspace{-0.198cm}
\begin{figure}[htb]
\begin{center}
\includegraphics[width=0.4898\textwidth]{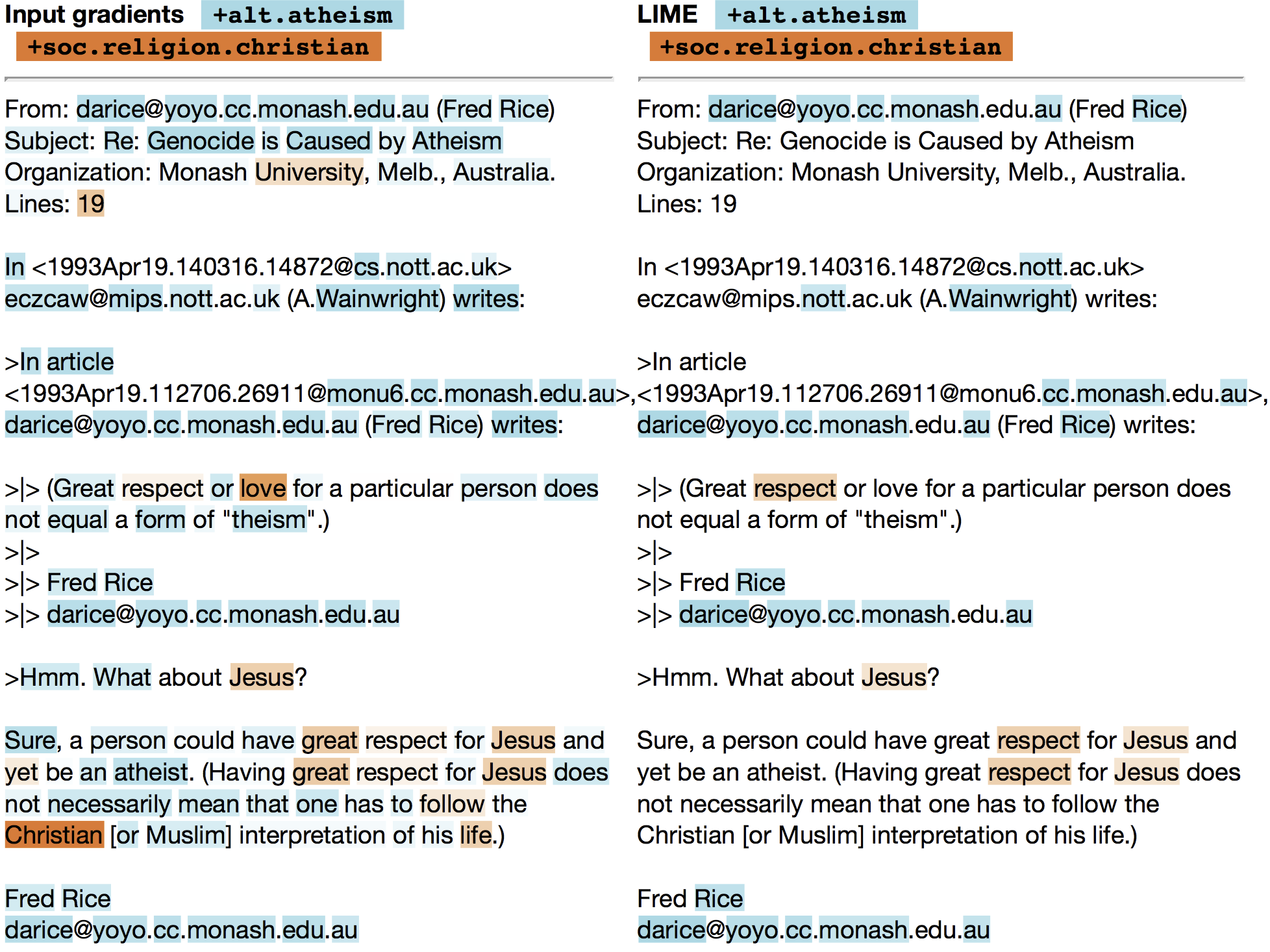} \\
\vspace{0.2cm}
\includegraphics[width=0.4898\textwidth]{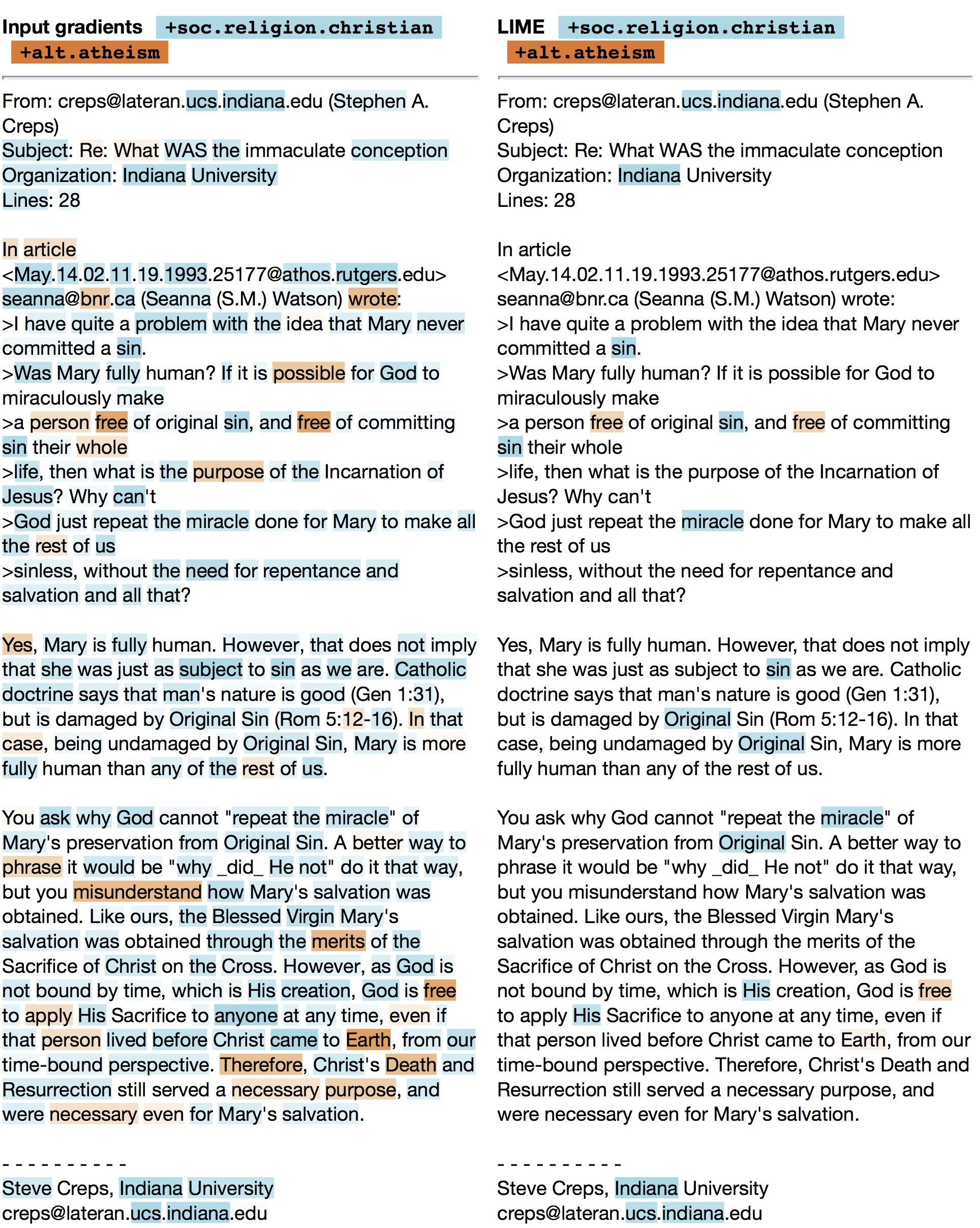}
\end{center}
\caption{\small Longer 20 Newsgroups examples. Blue supports the predicted label, orange opposes it, and $\text{opacity}_i = |w_i|/\max|w|$. LIME and input gradients never disagree, but gradients may provide a fuller picture of the model's behavior because of LIME's limits on features and samples (especially for long documents).}
\label{fig:longer-20ng}
\end{figure}